\documentclass[times,onecolumn,preprint,authoryear]{elsarticle}

\usepackage{prletters}
\usepackage{framed,multirow}

\usepackage{amssymb}
\usepackage{latexsym}
\usepackage{setspace}
\usepackage{ulem}
\usepackage{subfigure}
\usepackage{hyperref}
\usepackage{graphicx}
\usepackage{wrapfig}
\usepackage{caption}
\usepackage{pifont}

\journal{Pattern Recognition Letters}
\spacing{0.25}
\begin{document}

\begin{frontmatter}

\title{Understanding Trained CNNs by Indexing Neuron Selectivity}

\author[1]{Ivet Rafegas} 
\author[1]{Maria Vanrell\corref{cor1}}
\cortext[cor1]{Corresponding author: 
  Tel: +34 93 581 2415;
  Fax: +34 93 581 1670;
}
\ead{maria.vanrell@uab.cat}
\author[2]{Lu\'{\i}s A. Alexandre}
\author[1]{Guillem Arias}

\address[1]{Computer Vision Center. Universitat Autònoma de Barcelona. Edifici O, Campus UAB-Bellaterra (Barcelona), Spain}
\address[2]{ Universidade da Beira Interior and  Instituto de Telecomunica\c{c}\~oes, Rua Marqu\^es d'\'Avila e Bolama, 6201-001, Covilh\~a, Portugal}

\begin{abstract}
The impressive performance of Convolutional Neural Networks (CNNs) when solving different vision problems is shadowed by their black-box nature and our consequent lack of understanding of the representations they build and how these representations are organized. To help understanding these issues, we propose to describe the activity of individual neurons by their Neuron Feature visualization and quantify their inherent selectivity with two specific properties. We explore selectivity indexes for: an image feature (color); and an image label (class membership). Our contribution is a framework to seek or classify neurons by indexing on these selectivity properties. It helps to find color selective neurons, such as a \textit{red-mushroom neuron} in layer Conv4 or class selective neurons such as \textit{dog-face neurons} in layer Conv5 in VGG-M, and establishes a methodology to derive other selectivity properties. Indexing on neuron selectivity can statistically draw how features and classes are represented through layers in a moment when the size of trained nets is growing and automatic tools to index neurons can be helpful. 
\end{abstract}


\end{frontmatter}

\section{Introduction}\label{sec:introduction}

In parallel with the success of CNNs to solve vision problems, there has been a continual interest in developing methodologies to understand and visualize internal representations of these networks. How the responses of a trained CNN encode the visual information is a fundamental question for computer vision.

Several works have proposed different methodologies to address the understanding problem, in \cite{YixuanLi2015} two main groups of works were mentioned. On one side, those works that deal with the problem from a \textit{theoretical} point of view. These are works such as \cite{Montavon11} where kernel sequences are used to conclude that deep networks create increasingly better representations as the number of layer increases, \cite{Arnab14} which explains why a deep learning network learns simple features first and that the representation complexity increases as the layers get deeper, \cite{Goodfellow14} where an explanation for why an adversarial example created for one network is still valid in many others and they usually assign it the same (wrong) class. More recently, an emerging direction is building decision trees synthesizing the network generalization abilities \cite{FrosstHinton2017}.

\begin{table*}[!b]
\centering
\caption{Comparison of methods for visualizing internal CNN representations. Methods are classified according to their focus on: (a) individual neuron activity, (b) inverting the feature maps, or (c) modifying inputs or activations. The comparison is made taking into account the following aspects: (a) image-specific independence of visualization; (b) realistic appearance of visualization; and (c) ability of the method to characterize multiple properties of individual neurons.}\label{tab:taxonomy_methods}
\begin{tabular}{l|l|l|c|c|c|}
\shortstack[l]{Focus on} & Method goal & Reference & \shortstack{image\\independence} & \shortstack{realistic\\appearance} & \shortstack{multiple \\ properties}  \\ \hline

\multirow{5}{*}{\shortstack{Neuron \\ activity}} 
& visualizing weights & {[}Li \textit{et-al},16{]}       & yes &  yes  & no \\ \cline{2-6}
& projecting weights  & {[}Rafegas \textit{et-al},16{]}  & yes &  yes  & no \\ \cline{2-6}
& \multirow{3}{*}{\shortstack[l]{exploring intrinsic\\ features}} & {[}Girshick \textit{et-al},14{]}     & no     & yes & no \\ \cline{3-6}
&                                                                 & {[}Zeiler \textit{et-al},14{]}       & no     & yes & no \\ \cline{3-6}
&                                                                 & {[}Springenberg \textit{et-al},14{]} & no     & yes & no \\ \cline{2-6}
& \multirow{4}{*}{\shortstack[l]{generating image\\maximizing activation}} & {[}Simonyan \textit{et-al},14{]}  & yes & no     & no \\ \cline{3-6}
&                                                                 & {[}Yosinsky \textit{et-al},15{]}       & yes & no & yes \\ \cline{3-6}
&                                                                 & {[}Nguyen \textit{et-al},16a{]}      & yes & yes & no \\ \cline{3-6}
&                                                                 & {[}Nguyen \textit{et-al},16b{]}      & yes & no & yes \\ \cline{3-6}
&                                                                 & {[}Olah \textit{et-al},17-18{]}      & yes & no & yes \\ \cline{2-6}
& \multirow{2}{*}{\shortstack[l]{labelling \\semantic concepts}}   & {[}Zhou \textit{et-al},18{]},{[}Bau \textit{et-al},19{]} & no  & yes & no \\ \cline{3-6} 
&                                                         & {[}Fong \textit{et-al},18{]} & no & yes & yes \\ \cline{2-6}
& \shortstack[l]{averaging weighted\\top-scoring patches} & Our approach                    & yes & yes & yes \\ \hline
\multirow{2}{*}{\shortstack[l]{Inverting \\feature maps}} 
& generating & {[}Mahendran \textit{et-al},15{]}    & no  & yes & no \\[0.25cm] \cline{2-6}
& training   & {[}Dosovistkiy \textit{et-al},15{]}  & no  & yes & no \\[0.25cm] \hline
\multirow{3}{*}{Modifications}
& modifying activations & {[}Dosovitskiy \textit{et-al},16{]}  & no & yes & no \\ \cline{2-6}
& modifying stimuli     & {[}Aubry \textit{et-al},15{]}        & no & yes & no \\ \cline{2-6}
& ablation              & {[}Zhou  \textit{et-al},18{]}        & no & yes & no \\ \hline
\end{tabular}
\end{table*}

On the other side, an \textit{empirical} point of view, which comprises approaches that pursuit methodologies to visualize intermediate features in the image space, or approaches that analyze the effect of modifying a given feature map in a neuron activation.  One of the most relevant works pursuing the visualization of intermediate features, was the one proposed by \cite{Zeiler14}, where they projected intermediate image mappings into the image space using the deconvolution approach presented in \cite{Zeiler10}. By observing these projections that maximally activate a certain neuron they got the intuition about the main features learned on the network. Later on, in \cite{Dosovitskiy15a} the guided backpropagation improved the deconvolution approach through a new way of inverting rectified linear (ReLu) non-linearities, achieving better visualizations of the activations. Other works made use of optimization techniques to deal with the neuron visualization problem. The key point of these works was using an appropriate regularization in the process. \cite{Simonyan14} proposed a method to generate an image which is representative of a certain class by maximizing the score of this image to be classified in a certain class (or highly activates the specified neuron) with an $L_2$-regularization.  Afterwards,  \cite{Mahendran15} searched for an image whose feature map best matched a given feature map by incorporating natural image priors. A similar work was performed by \cite{Yosinski15} but taking advantage of combining three different regularizations to achieve more recognizable images. \cite{Dosovitskiy16} explored the internal representation of a given convolutional network by training a new deconvolutional network to learn filter weights that minimized the image reconstruction error when these filters were applied to the image feature maps. \cite{Nguyen_pref} used a trained deep generator network to obtain images that maximally activated neurons obtaining visually realistic results, while in its work \cite{Nguyen_mult} explored the possibility of creating multiple faceted images that encode multiple features activating the same neuron. In a similar line of work, \cite{olah2017feature} created a method to visualize neurons by optimizing images to produce high neuron activation, and combined these representations to visualize the possible interactions between different neurons and in \cite{olah2018the} they further explored how images are seen through a CNN, which neurons are activated in each region of a given image, and what semantic categories are related to those neurons. In recent studies, neuron activity visualization is shifting towards semantic characterization. \cite{Torralba2018_article} and \cite{Torralba2018_gan} studied the activation of the neurons for a dataset of multilabeled images with different concepts such as color, objects or materials, obtaining which labels of the images were more activated in each neuron and giving that neuron the correspondent semantic category in CNNs and GANs respectively. Following a similar methodology, \cite{Vedaldi_net2vec} studied how concepts are encoded by multiple neurons of the network rather than individual neurons.

In the second subset of empirical approaches, \cite{Dosovitskiy15a} trained a generative deconvolutional network to create images from neuron activations. With this methodology, the variation of the activations enables the visualization of the differences in the generated images. A similar analysis is done by \cite{Aubry15}, but instead of backpropagating different activations to the image space and comparing them, they observed the changes on neuron activations when similar computer-generated images with different scene factors were introduced into a CNN. Both works concluded that there were specific neurons sensitive to color changes, point of views, scale or lighting configurations. In a more simplistic approach, \cite{torralba_ablation} uses ablation to remove neurons from the network and looks at the effect of these ablations in the performance of the network.

In this work we propose a framework to understand internal representations of CNNs, by associating selectivity indexes to individual neurons. Firstly, we propose to work on the set of images that maximally activates a neuron, and from these images we visualize its \textit{neuron feature} that usually outlines the intrinsic feature that spikes the neuron. Afterwards, we propose two selectivity indexes which are different in their essence: a \textit{color selectivity index} that quantifies the degree of response of a  neuron to a specific color and a \textit{class selectivity index} that quantifies the degree of response of a neuron to a specific class label. Indexes are derived from the neuron features or directly from the set of images with maximum activations. 

We analyze both indexes on a VGG-M network (\cite{Chatfield14}) trained on ImageNet (\cite{imagenet_cvpr09}) and we confirm their flexibility to cluster neurons according to their index values and extract conclusions in terms of their task in the net. By analyzing color selective neurons we are able to outline how color is represented by the network. Curiously, some parallelisms were found between color representation in the internal representation of a CNN and the human visual system \cite{rafegas2018,rafegas2017}. Color selective-neurons also show some preferences towards specific colors which coincide with ImageNet color biases. Indexing on class selectivity we found highly class selective neurons like \textit{digital-clock} at conv2, \textit{cardoon} at conv3 and \textit{ladybug} at conv5, much before the fully connected layers.

To sum up, in table \ref{tab:taxonomy_methods} we propose a taxonomy for the revised approaches dealing with the problem of understanding and visualizing CNN internal representations. The works have been classified in three groups according to their objective: (a) characterizing individual neuron activity; (b) inverting the feature maps; and (c) applying controlled modifications either to the input or to the activations. For each of the approaches we analyze three different properties of the proposed methodology.

\section{Neuron Feature}\label{sec:neuronfeature}
In this paper we propose to visualize the image features that activate a neuron by directly computing a weighted average of the $N$-th first images that maximally activate this neuron. We term it as the \textit{Neuron Feature} (NF) and it is an image independent visualization of the type of data that most activates a single neuron. 

\begin{figure*}[!b]
\begin{center}
 \includegraphics[width=16.2cm]{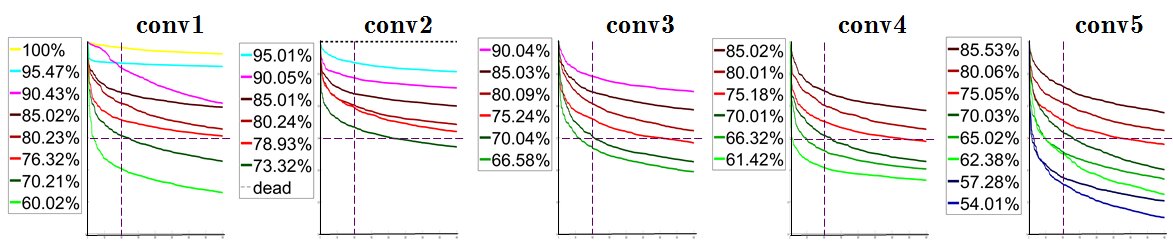}
  \end{center}
  \caption{Ranked activation of a subset of neurons for the different convolutional layers for the 400 first ranked images (compared through all image dataset). Each figure shows the highest and smallest area of activation achieved for each layer, and some examples between these extremes. }\label{fig:activations}
\end{figure*}

\begin{figure*}[!b]
\begin{center}
\includegraphics[width=16.2cm]{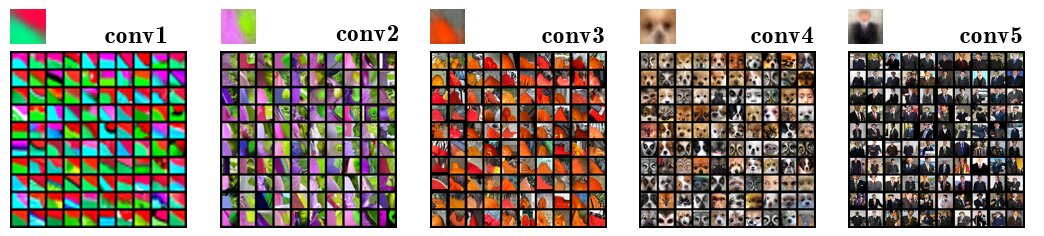}
  \end{center}
  \caption{Visualization of 5 NF for different layers of VGG-M with their corresponding 100 cropped images, receptive fields are all scaled to the same size.}\label{fig:nf_and_cropped}
\end{figure*}

In order to build the NF we need to calculate the activations associated to each individual neuron. For each neuron we select the set of images that achieve a minimum relative activation value but constrained to a maximum number of images, $N_{max}$, for practical reasons. In Fig.~\ref{fig:activations} we can see the behavior of the ranked responses of a subset of neurons for every convolution layer of the VGG-M CNN trained on ImageNet by \cite{Chatfield14}. We can see neurons with different behaviors, from neurons that activate for almost all the images, those with a higher area under the activation curve, to more specialized neurons that just activate for a subset of images, those with a smaller activation area, with the highest activations concentrated in the first ranked images. The observation of the responses confirms the adequacy of our assumption to fix a minimum value for the activation and a maximum number of images to capture the most important activations for all the neurons. Similar observations have been made for other networks like VGG-S and VGG-F \cite{Chatfield14}\footnote{The results are shown for a maximum number of images equal to $N_{max}=100$ and a minimum activation value over a $70\%$ of the maximum activation. We plot these values on Fig. ~\ref{fig:activations}}. Thus, the NF is computed as: 

\begin{equation}
NF(n^{L,i}) = \frac{1}{N_{max}} \sum_{j=1}^{N_{max}}w_{j,i,L} I_j
\end{equation}
where $w_{j,i,L}$ is the relative activation of the $j$-th cropped image, denoted as $I_j$, of the $i$-th neuron $n^{L,i}$ at layer $L$. The relative activation is the activation $a_{j,i}$ of a neuron, given an input image,  with respect to its maximum activation obtained for any image,  $w_{j,i,L}=\frac{a_{j,i}}{a_{max,i}}$ where $a_{max,i} =\max a_{k,i}, \forall k$.

\begin{figure}[t]
\begin{center}
\includegraphics[width=8.1cm]{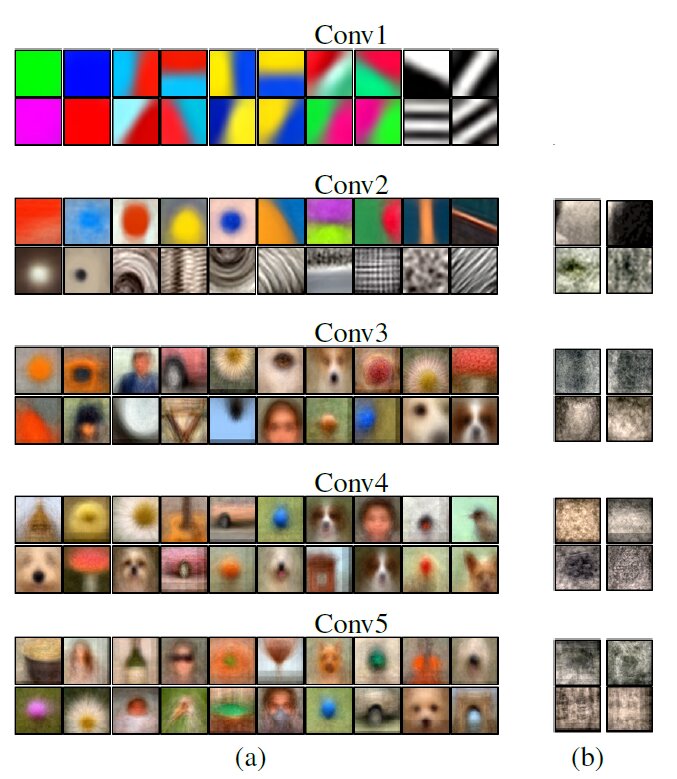}
\end{center}
\caption{Examples of NFs for each convolutional layer of the network VGG-M. (a) 20 examples per convolutional layer of well defined NF; (b) examples of blurred NF. Although sizes of NF increase with layer depth, we scale them into the same size. Original sizes are: 7x7 , 27x27, 75x75, 107x107 and 139x139.}\label{fig:exemples_NF}
\end{figure}

In Fig.~\ref{fig:nf_and_cropped} we can see some NFs and their corresponding set of first $100$ image crops that produce maximum activations, and in Fig.~\ref{fig:exemples_NF} (a) we can see a selected subset of 20 NF per layer. In this image we can identify specific shapes that display the intrinsic property that fires a single neuron. At first glance, we can see how in this particular network the first two layers are devoted to basic properties. Oriented edges of different frequencies and in different colors in the first layer; textures, blobs, bars and more specific curves in the second layer. The rest of the layers seem to be devoted to more complex objects. We can see that dog and human faces, cars and flowers are detected at different scales in different layers, since the size of the NF and their corresponding cropped images increase with depth. However, not all neurons present such a clear tuning to an identifiable shape. Some neurons present a blurred version of NF, such as, those in Fig.~\ref{fig:exemples_NF}(b). The level of blurring is directly related to a high variability between the crops that maximally activate the neuron. 

At this point, we want to make a short parenthesis to relate the previous representational observations with the neural coding issue that is the focus of attention in visual brain research (\cite{Kriegeskorte11}). We are referring to the hypothesis about distributed representations that encode object information in neuron population codes, that co-exist with strong evidences of neurons which are only activated by a very specific object. In line with this idea, we speculate whether neurons presenting a highly structured NF could be closer to localist code neurons while neurons with a blurred NF could be closer to a distributed code. We return to this discussion later on in sections \ref{subsec:class_select} and \ref{sec:conclusions}.




\section{Neuron Selectivity Indexes}

In this section we propose to describe neurons by their inherent response to a specific property, using an index. The index has to allow the ranking of the neurons based on its response and the existence of the property in the input image. Therefore, we translate the problem of describing neuron activity to the problem of proposing methods which are able to quantify specific image facets that  correlate with the degree of activation of the neuron holding such a property. A selectivity index of a single neuron is a flexible and independent method for discriminating or clustering neurons inside the same network.  Selectivity indexes can be defined either for image features or for image labels, in what follows, we propose two selectivity indexes, one in each group.

\subsection{Color Selectivity Index} 
Color selectivity is a property that has been measured in the human brain. The level of activation of specific neurons when the observer is exposed to a stimulus with a strong color bias, and its corresponding low activation when the color is not present, is the object of attention in vision research that pursuits the understanding of how color is coded in the human visual system (\cite{Shapley11}). 

Inspired by this, here we propose a method to compute a color selectivity index for neurons in artificial neural networks. We propose to base it directly on the image properties of the NF we have defined above. We quantify the selectivity to a specific chromaticity directly from the color distribution of the NF. We define this index as the angle between the first principal component ($\mathbf{v}$)  of the color distribution of the NF and the intensity axis ($\mathbf{b}$) of the Opponent Color Space (OPP). To compute ($\mathbf{v}$) we use a weighted Principal Component Analysis \cite{Delchambre14} that allows to strengthen the selectivity of small color areas. Weights are applied to each pixel in order to reinforce those pixels that are shared by most cropped images and that highly contribute to the NF. Therefore, the weights are the inverse of the standard deviation. In this way, a NF defined by cropped images with different colors will tend to be represented by a grayish image and its principal component will be close to the intensity axis in the OPP color space and it will receive a low selectivity index. We formulate this index (in degrees) as follows:
\begin{equation}
\alpha \left(n^{L,i}\right) = \frac{1}{90} \arccos \left(\frac{\mathbf{b} \cdot \mathbf{v} }{\| \mathbf{b} \| \  \|\mathbf{v}\| } \right) 
\end{equation}
 
Other selectivity indexes that can be derived from this, are those related to color attributes. We can easily extract color name labels using a color naming approach such as \cite{Benavente08} and directly define color selectivity to basic names such as red, or green, among others. 

\subsection{Class selectivity index} 

Class selectivity is a property of a neuron that can help to establish its discriminative power for one specific class or can allow to cluster neurons according to the ontological properties of their class labels. 

We propose a method to compute a class selectivity index for individual neurons by compiling the class labels of the images that maximally activate this neuron, in a single descriptor. We define class selectivity from the set of class labels of the $N$ images used to build the NF. To quantify this index we build the class label distribution of the full set of images. As in the color selectivity index, we weight the significance of a class label by the relative activation of its image. Thus, the relative frequency of each class $c$ for a certain neuron is defined as: 
\begin{equation}
f_c\left(n^{i,L}\right) =\frac{\sum_j^{N_{c}} w_{j,i,L}}{\sum_l^{N} w_{l,i,L}}
\end{equation}
where $N_c$ refers to the number of images, among the $N$ cropped images activating this neuron, that belong to class $c$.

Finally, our class selectivity index is defined as follows: 
\begin{equation}
\gamma \left(n^{L,i}\right)=\frac{N - M}{N-1}
\end{equation}
where $M$ is the minimum number of classes that covers a pre-fixed ratio, $th$, of the neuron activation \textit{i.e.},
$\sum_{c=1}^M f_c \geq th$. 
This threshold allows us to avoid considering class labels with very small activation weights. In this way, we get a global class selectivity index plus a set of $M$ classes that describes what classes the neuron is selective to. Therefore, a high value (maximum is 1) indicates a strong contribution of this neuron to a single class, while a low class selectivity index indicates a poor contribution of this neuron to a single class (minimum is 0 when $M=N$). In between we can have different degrees of selectivity to different number of classes. 

This index can also contribute to give some insights about the problem of how information is coded through layers, in the debate of localist and distributed neural codes we mentioned before (\cite{Kriegeskorte11}). Neurons with high class selectivity index should be in line with a localist code, while neurons with low class selectivity index should be part of a distributed code. The way the index is defined allows a large range of interpretations in between these two kinds of coding as it has been outlined in the visual coding literature.

\section{Results}
In this section we report some empirical results to show how the proposed selectivity indexes perform and what representational conclusions we can extract from the subsets of neurons sharing indexed properties. 

\subsection{Experimental setup}
\label{sec:experimental_setup}
We analyze the neurons of a CNN architecture trained on ImageNet ILSVRC dataset \cite{imagenet_cvpr09} (using a subset of 1.2M images classified in 1.000 categories). We report the results for the VGG-M CNN that was trained by \cite{Chatfield14} for a generic visual task of object recognition. This CNN consists of 5 convolutional layers of 96, 256, 512, 512 and 512 neurons respectively, followed by two fully connected layers of 4096 units each and a final classification layer of 1000 units. We selected this network since it has a similar structure to those which have been reported as having a representational performance that competes with human performance (as was proved in \cite{dicarlo14}). Nevertheless, we have obtained similar results for VGG-F and VGG-S that are provided  in \cite{Chatfield14}. We used the Matconvnet library provided by \cite{vedaldi15matconvnet} for all the experiments.

\subsection{Color selectivity}

General purpose CNN architectures are usually trained on RGB color images. However there is a strong belief in the computer vision community that color is a dispensable property. The results we obtain by indexing color selective neurons make us conclude that there is no basis for such a belief. Results show that color is strongly entangled at all levels of the CNN representation. In a preliminary experiment we have tested a subset of ImageNet images with VGG-M in their original color and the same subset in a gray scale representation. Classification results show a considerable decrease: while original RGB images are classified with a 27.50\% top-1 error and 10.14\% top-5 error, gray scale image versions present 51.12\% and 26.37\% errors, top-1 and top-5 errors respectively. 

\begin{figure*}
\centering
   \includegraphics[width=17cm]{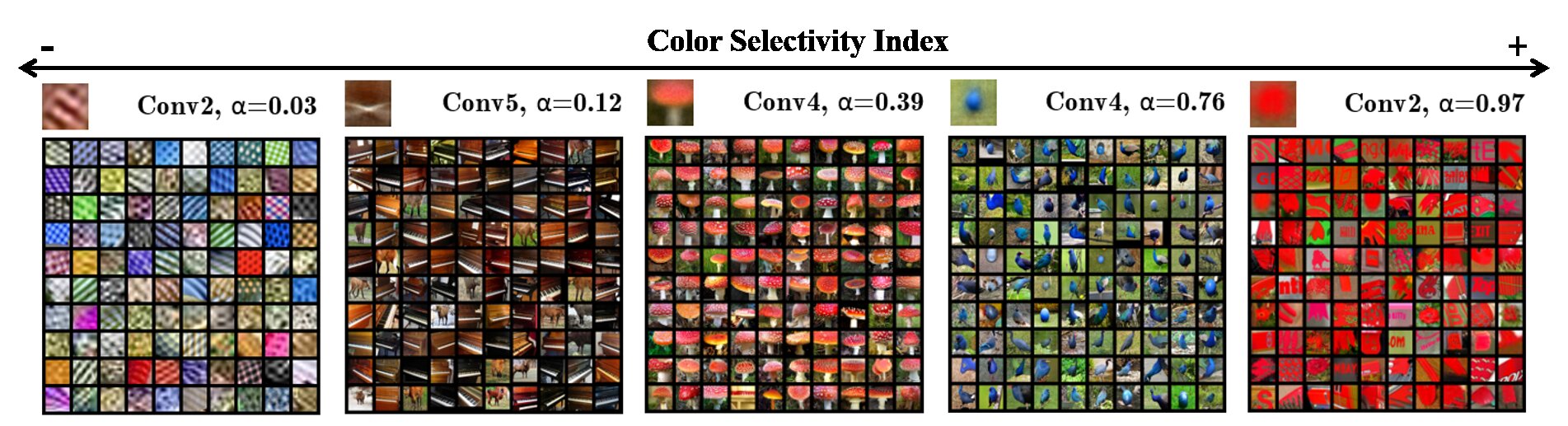}
   \caption{Neurons with different color selectivity indexes. For each neuron, the NF (top) and its cropped images (bottom).  
   }\label{fig:ex_color_selectivity}
\end{figure*}

\begin{figure}[!t]
  \centering
  \begin{minipage}[b]{0.45\textwidth}
    \includegraphics[width=7.5cm]{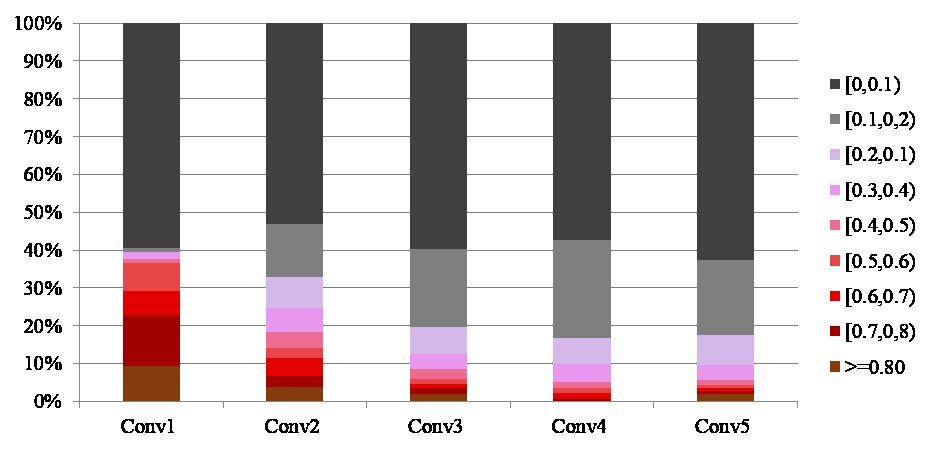}
    \caption{Number and index of color selectivity neurons through layers. Grayish and reddish bars are for low and high index values, respectively.}\label{fig:graphical_color}
    \begin{center}
    \end{center}
    \end{minipage}
  \hfill
  \begin{minipage}[b]{0.45\textwidth}
    \includegraphics[width=7.5cm]{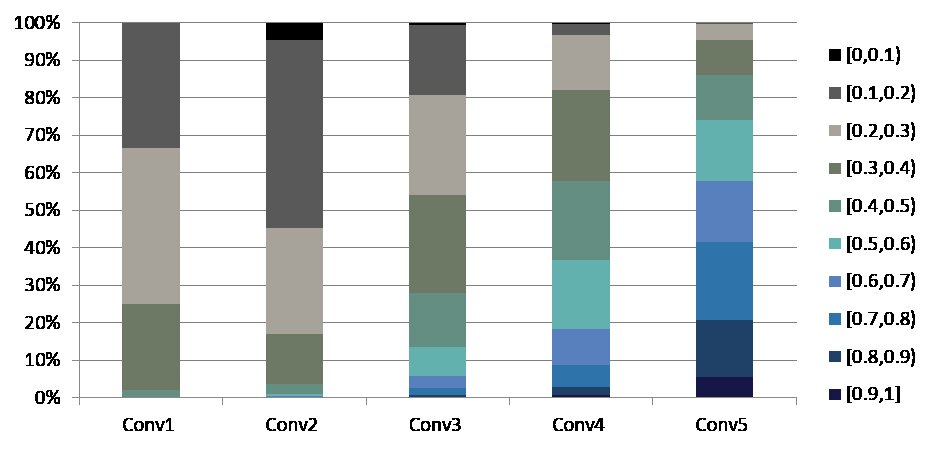}
    \caption{Number and index of class selectivity neurons through layers. Grayish and bluish bars are for low and high index values, respectively. }\label{fig:graphical_class}
    \begin{center}
    \end{center}
  \end{minipage} 
\end{figure}

We extracted how many neurons are related to color in each convolutional layer using the proposed color selectivity index. The bars in Fig.~\ref{fig:graphical_color} plot the relative number of neurons that are color selective compared to those that are not. Grey represents the ratio of neurons that do not spike for the presence of a color and reddish represent neurons that are highly activated by the presence of a color. In the graphic we can observe that shallow layers are the main responsible for the color representation on the images: $40\%$ and $30\%$ of neurons are color selective in layers conv1 and conv2, respectively. Nevertheless, we also still found around $20\%$ of color selective neurons in deeper layers. Therefore, although neurons in deeper layers tend to be color invariant, an important part of the network's representation capability is devoted to color. This fact, supports the importance of color in object recognition. In Fig.~\ref{fig:ex_color_selectivity} we show some examples of NFs with different degrees of color selectivity on different layers of the network and showing the corresponding cropped images.

\subsection{Class selectivity}\label{subsec:class_select}
Following with the analysis of ranking neurons by their response to a certain property, here we focus on the proposed selectivity index that relates to image labels instead of to an image property, the class selectivity index, which only applies for classification networks. We report the results of different experiments where we have fixed $th=1$, which means we consider all the class labels for the $N=100$ images that maximally activates the neuron. As we mentioned before, this index can enlighten how classes are encoded through the net layers, which  can be related to the problem of how objects are encoded in the human brain. Here we hypothesize that the difference between localist or distributed codes could correlate with the idea of neurons highly selective to a single class and neurons highly selective to several classes. We resume on this topic later in section \ref{sec:conclusions}.
In a first experiment on class selectivity, we analyze how many neurons present different degrees of class selectivity through layers. The bars in Fig.~\ref{fig:graphical_class} plot the relative quantity of neurons that are class selective compared to those that are not. Grey represents the ratio of neurons that are not activated by a single class and bluish represents neurons that are highly activated by a single class. Opposite to what we showed about color selectivity, we found most of class selective neurons in deeper layers, and no class selectivity in shallow layers, as expected. We have moved from a very basic image property, color, to a very high level property, class label. This fact corroborates the idea that CNNs start by defining basic feature detectors that are shared by most of the classes, and the neurons become more specialized when they belong to deeper layers representing larger areas in the image space and therefore more complex shapes. We start to have neurons with relevant class selectivity in layer conv3, where a $5\%$ of the neurons are quite class selective and we found some neurons with a degree of selectivity close to $1$. These ratios progressively increase up to layer conv5 where we have more than $50\%$ of neurons with a class selectivity index greater than $0.6$, that means that we have less than 40 different classes activating each of these neurons, which is a very selective ratio considering the number of classes of the ImageNet dataset. In the same layer $20\%$ of neurons present a high class selectivity index, meaning that they are activated by less than 20 different classes. 

\begin{figure*}
\centering
   \includegraphics[width=17cm]{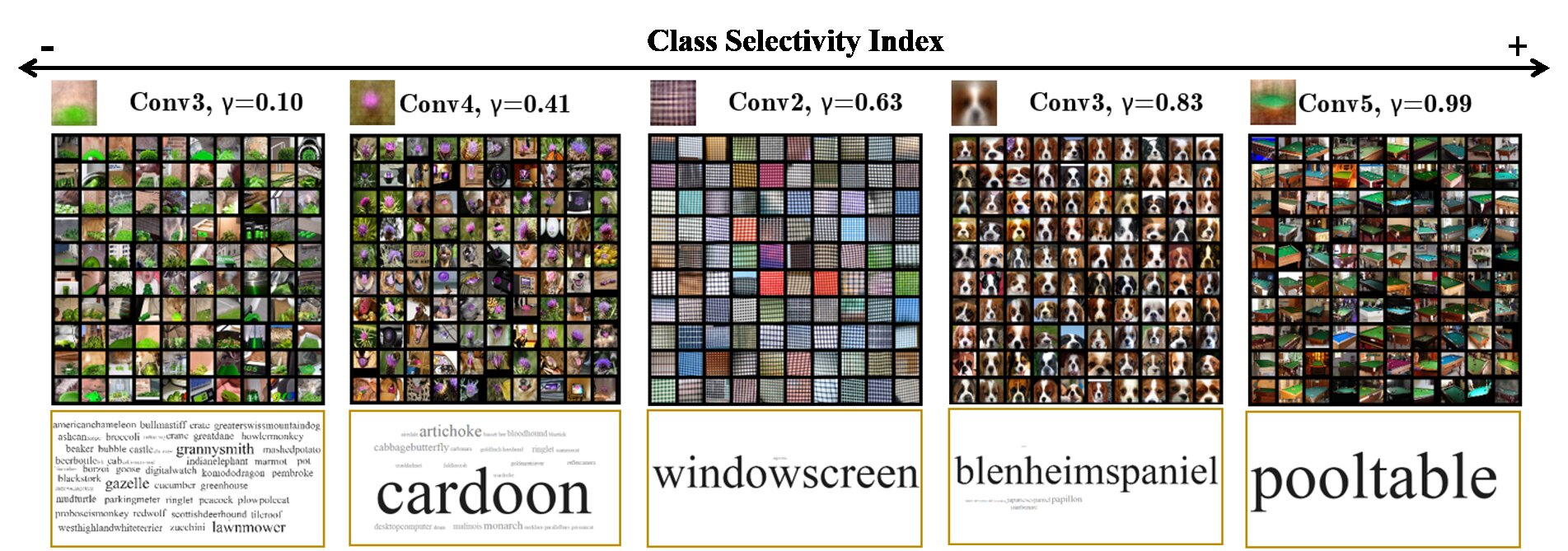}
   \caption{Neurons with different class selectivity indexes. For each neuron, the NF, its cropped images and a tag cloud of classes in the Wordnet ontology.    }\label{fig:ex_class_selectivity}
\end{figure*}

Secondly, we have visualized the properties of a set of images presenting different degrees of class selectivity in  Fig.~\ref{fig:ex_class_selectivity} for different levels of depth. We represent each neuron with their NF visualization and the corresponding cropped images. We also show two \textit{tag clouds} of each neuron. They show the importance of each class label. With an orange frame we plot the leaf classes of the ImageNet ontology. This second analysis could help finding neurons that are specialized on a general semantic concept that different classes share. Note that neurons with high class selectivity index have a set of cropped images that we can identify as belonging to the same class. 

\section{Conclusions}\label{sec:conclusions}

In this paper we proposed a framework to analyze a trained CNN by dissecting individual neurons using their indexes of selectivity to specific properties. We  have proposed two indexes based on properties of different nature: (a) color, that is a low-level image property that we have shown to be entangled in all the representations levels of the net; (b) class label, that is a high-level image property that can be analyzed at different levels of abstraction. We have shown that while the number of color selective neurons decreases with depth, the number of class selective neuron increases. We have also proposed to visualize the image structures that cause high activity of a neuron, with what we have called the neuron feature (NF).

The insights gained from the proposed work have made us speculate about the two different ways to address the coding properties of individual neurons (localist versus distributed). Firstly, we have mentioned the possibility that a blurred NF, i.e., without a clear structure, belongs to a neuron that can be part of a distributed code where the neuron is not selective to a single shape, but maybe to diverse shapes than can be part of a code in deeper neurons. Secondly, we speculated about the possibility that neurons with high class selectivity index can be part of a localist code, whereas those that have a low index would be part of a distributed representation. In parallel, the analysis of the color selective neurons indicated some parallelism between color representation in the 1st conv layer and known evidences about the color representation in the human visual system. 

As further work we intend to fully exploit the potential of the indexes in different neural network architectures, defining new selectivity indexes for shape, texture or part detection, that could be a perfect complement to current ones, and study the relations between them, for instance, what classes are related to certain colors or textures.

\vspace{0.25cm}
\noindent \textbf{Acknowledgments.} This work has been partially funded by project TIN2014-61068-R MINECO. 3rd author was partially funded by FCT/MEC through national funds and when applicable co-funded by FEDER - PT2020 partnership agreement under the project UID/EEA/50008/2019.

\bibliographystyle{model2-names}
\bibliography{refs}

\begin{thebibliography}{31}
\expandafter\ifx\csname natexlab\endcsname\relax\def\natexlab#1{#1}\fi
\providecommand{\url}[1]{\texttt{#1}}
\providecommand{\href}[2]{#2}
\providecommand{\path}[1]{#1}
\providecommand{\DOIprefix}{doi:}
\providecommand{\ArXivprefix}{arXiv:}
\providecommand{\URLprefix}{URL: }
\providecommand{\Pubmedprefix}{pmid:}
\providecommand{\doi}[1]{\href{http://dx.doi.org/#1}{\path{#1}}}
\providecommand{\Pubmed}[1]{\href{pmid:#1}{\path{#1}}}
\providecommand{\bibinfo}[2]{#2}
\ifx\xfnm\relax \def\xfnm[#1]{\unskip,\space#1}\fi
\bibitem[{Aubry and Russell(2015)}]{Aubry15}
\bibinfo{author}{Aubry, M.}, \bibinfo{author}{Russell, B.C.},
  \bibinfo{year}{2015}.
\newblock \bibinfo{title}{Understanding deep features with computer-generated
  imagery}, in: \bibinfo{booktitle}{Proc. of ICCV}.
\bibitem[{Bau et~al.(2018)Bau, Zhu, Strobelt, Zhou, Tenenbaum, Freeman and
  Torralba}]{Torralba2018_gan}
\bibinfo{author}{Bau, D.}, \bibinfo{author}{Zhu, J.},
  \bibinfo{author}{Strobelt, H.}, \bibinfo{author}{Zhou, B.},
  \bibinfo{author}{Tenenbaum, J.B.}, \bibinfo{author}{Freeman, W.T.},
  \bibinfo{author}{Torralba, A.}, \bibinfo{year}{2018}.
\newblock \bibinfo{title}{{GAN} dissection: Visualizing and understanding
  generative adversarial networks}.
\newblock \bibinfo{journal}{CoRR} \URLprefix
  \url{https://arxiv.org/abs/1901.09887}.
\bibitem[{Benavente et~al.(2008)Benavente, Vanrell and Baldrich}]{Benavente08}
\bibinfo{author}{Benavente, R.}, \bibinfo{author}{Vanrell, M.},
  \bibinfo{author}{Baldrich, R.}, \bibinfo{year}{2008}.
\newblock \bibinfo{title}{Parametric fuzzy sets for automatic color naming}.
\newblock \bibinfo{journal}{JOSA} \bibinfo{volume}{25},
  \bibinfo{pages}{2582--2593}.
\bibitem[{Cadieu et~al.(2014)Cadieu, Hong, Yamins, Pinto, Ardila, Solomon,
  Majaj and {DiCarlo}}]{dicarlo14}
\bibinfo{author}{Cadieu, C.F.}, \bibinfo{author}{Hong, H.},
  \bibinfo{author}{Yamins, D.L.K.}, \bibinfo{author}{Pinto, N.},
  \bibinfo{author}{Ardila, D.}, \bibinfo{author}{Solomon, E.A.},
  \bibinfo{author}{Majaj, N.J.}, \bibinfo{author}{{DiCarlo}, J.J.},
  \bibinfo{year}{2014}.
\newblock \bibinfo{title}{Deep neural networks rival the representation of
  primate it cortex for core visual object recognition.}
\newblock \bibinfo{journal}{PLoS computational biology} \bibinfo{volume}{10},
  \bibinfo{pages}{1--18}.
\bibitem[{Chatfield et~al.(2014)Chatfield, Simonyan, Vedaldi and
  Zisserman}]{Chatfield14}
\bibinfo{author}{Chatfield, K.}, \bibinfo{author}{Simonyan, K.},
  \bibinfo{author}{Vedaldi, A.}, \bibinfo{author}{Zisserman, A.},
  \bibinfo{year}{2014}.
\newblock \bibinfo{title}{Return of the devil in the details: Delving deep into
  convolutional nets}.
\newblock \bibinfo{journal}{Proc. of BMVC} .
\bibitem[{Delchambre(2014)}]{Delchambre14}
\bibinfo{author}{Delchambre, L.}, \bibinfo{year}{2014}.
\newblock \bibinfo{title}{Weighted principal component analysis: a weighted
  covariance eigendecomposition approach}.
\newblock \bibinfo{journal}{Monthly Notices of the Royal Astronomical Society}
  \bibinfo{volume}{446}, \bibinfo{pages}{3545--3555}.
\bibitem[{Deng et~al.(2009)Deng, Dong, Socher, Li, Li and
  Fei-Fei}]{imagenet_cvpr09}
\bibinfo{author}{Deng, J.}, \bibinfo{author}{Dong, W.},
  \bibinfo{author}{Socher, R.}, \bibinfo{author}{Li, L.J.},
  \bibinfo{author}{Li, K.}, \bibinfo{author}{Fei-Fei, L.},
  \bibinfo{year}{2009}.
\newblock \bibinfo{title}{{ImageNet: A Large-Scale Hierarchical Image
  Database}}, in: \bibinfo{booktitle}{Proc. CVPR}.
\bibitem[{Dosovitskiy and Brox(2016)}]{Dosovitskiy16}
\bibinfo{author}{Dosovitskiy, A.}, \bibinfo{author}{Brox, T.},
  \bibinfo{year}{2016}.
\newblock \bibinfo{title}{Inverting visual representations with convolutional
  networks}.
\newblock \bibinfo{journal}{Proc. CVPR} .
\bibitem[{Dosovitskiy et~al.(2015)Dosovitskiy, Springenberg and
  Brox}]{Dosovitskiy15a}
\bibinfo{author}{Dosovitskiy, A.}, \bibinfo{author}{Springenberg, J.},
  \bibinfo{author}{Brox, T.}, \bibinfo{year}{2015}.
\newblock \bibinfo{title}{Learning to generate chairs with convolutional neural
  networks}, in: \bibinfo{booktitle}{Proc. CVPR}.
\bibitem[{Fong and Vedaldi(2018)}]{Vedaldi_net2vec}
\bibinfo{author}{Fong, R.}, \bibinfo{author}{Vedaldi, A.},
  \bibinfo{year}{2018}.
\newblock \bibinfo{title}{Net2vec: Quantifying and explaining how concepts are
  encoded by filters in deep neural networks}, in: \bibinfo{booktitle}{Proc.
  CVPR}.
\bibitem[{Frosst and Hinton(2017)}]{FrosstHinton2017}
\bibinfo{author}{Frosst, N.}, \bibinfo{author}{Hinton, G.},
  \bibinfo{year}{2017}.
\newblock \bibinfo{title}{Distilling a neural network into a soft decision
  tree}.
\newblock \bibinfo{journal}{CoRR} \URLprefix
  \url{http://arxiv.org/abs/1711.09784}.
\bibitem[{Goodfellow et~al.(2014)Goodfellow, Shlens and Szegedy}]{Goodfellow14}
\bibinfo{author}{Goodfellow, I.J.}, \bibinfo{author}{Shlens, J.},
  \bibinfo{author}{Szegedy, C.}, \bibinfo{year}{2014}.
\newblock \bibinfo{title}{Explaining and harnessing adversarial examples}.
\newblock \bibinfo{journal}{CoRR} \URLprefix
  \url{https://ui.adsabs.harvard.edu/abs/2014arXiv1412.6572G}.
\bibitem[{Kriegeskorte and Kreiman(2011)}]{Kriegeskorte11}
\bibinfo{author}{Kriegeskorte, N.}, \bibinfo{author}{Kreiman, G.},
  \bibinfo{year}{2011}.
\newblock \bibinfo{title}{Visual Population Codes - Toward a Common
  Multivariate Framework for Cell Recording and Functional Imaging}.
\newblock \bibinfo{publisher}{MIT Press}.
\bibitem[{Li et~al.(2016)Li, Yosinski, Clune, Lipson and
  Hopcroft}]{YixuanLi2015}
\bibinfo{author}{Li, Y.}, \bibinfo{author}{Yosinski, J.},
  \bibinfo{author}{Clune, J.}, \bibinfo{author}{Lipson, H.},
  \bibinfo{author}{Hopcroft, J.E.}, \bibinfo{year}{2016}.
\newblock \bibinfo{title}{Convergent learning: Do different neural networks
  learn the same representations?}, in: \bibinfo{booktitle}{Int. Conf. on
  Learning Representations}.
\bibitem[{Mahendran and Vedaldi(2015)}]{Mahendran15}
\bibinfo{author}{Mahendran, A.}, \bibinfo{author}{Vedaldi, A.},
  \bibinfo{year}{2015}.
\newblock \bibinfo{title}{Understanding deep image representations by inverting
  them}.
\newblock \bibinfo{journal}{Proc. CVPR} .
\bibitem[{Montavon et~al.(2011)Montavon, Braun and M\"{u}ller}]{Montavon11}
\bibinfo{author}{Montavon, G.}, \bibinfo{author}{Braun, M.},
  \bibinfo{author}{M\"{u}ller, K.R.}, \bibinfo{year}{2011}.
\newblock \bibinfo{title}{Kernel analysis of deep networks}.
\newblock \bibinfo{journal}{Journal of Machine Learning Research}
  \bibinfo{volume}{12}, \bibinfo{pages}{2563--2581}.
\bibitem[{Nguyen et~al.(2016a)Nguyen, Dosovitskiy, Yosinski, Brox and
  Clune}]{Nguyen_pref}
\bibinfo{author}{Nguyen, A.}, \bibinfo{author}{Dosovitskiy, A.},
  \bibinfo{author}{Yosinski, J.}, \bibinfo{author}{Brox, T.},
  \bibinfo{author}{Clune, J.}, \bibinfo{year}{2016}a.
\newblock \bibinfo{title}{Synthesizing the preferred inputs for neurons in
  neural networks via deep generator networks}, in:
  \bibinfo{booktitle}{NIPS'16}, pp. \bibinfo{pages}{3395--3403}.
\bibitem[{Nguyen et~al.(2016b)Nguyen, Yosinski and Clune}]{Nguyen_mult}
\bibinfo{author}{Nguyen, A.M.}, \bibinfo{author}{Yosinski, J.},
  \bibinfo{author}{Clune, J.}, \bibinfo{year}{2016}b.
\newblock \bibinfo{title}{Multifaceted feature visualization: Uncovering the
  different types of features learned by each neuron in deep neural networks}.
\newblock \bibinfo{journal}{Proc. of ICML Workshop} \bibinfo{volume}{48}.
\bibitem[{Olah et~al.(2017)Olah, Mordvintsev and Schubert}]{olah2017feature}
\bibinfo{author}{Olah, C.}, \bibinfo{author}{Mordvintsev, A.},
  \bibinfo{author}{Schubert, L.}, \bibinfo{year}{2017}.
\newblock \bibinfo{title}{Feature visualization}.
\newblock \bibinfo{journal}{Distill} \URLprefix
  \url{https://distill.pub/2017/feature-visualization/}.
\bibitem[{Olah et~al.(2018)Olah, Satyanarayan, Johnson, Carter, Schubert, Ye
  and Mordvintsev}]{olah2018the}
\bibinfo{author}{Olah, C.}, \bibinfo{author}{Satyanarayan, A.},
  \bibinfo{author}{Johnson, I.}, \bibinfo{author}{Carter, S.},
  \bibinfo{author}{Schubert, L.}, \bibinfo{author}{Ye, K.},
  \bibinfo{author}{Mordvintsev, A.}, \bibinfo{year}{2018}.
\newblock \bibinfo{title}{The building blocks of interpretability}.
\newblock \bibinfo{journal}{Distill} \URLprefix
  \url{https://distill.pub/2018/building-blocks/}.
\bibitem[{Paul and Venkatasubramanian(2014)}]{Arnab14}
\bibinfo{author}{Paul, A.}, \bibinfo{author}{Venkatasubramanian, S.},
  \bibinfo{year}{2014}.
\newblock \bibinfo{title}{Why does deep learning work? - {A} perspective from
  group theory}.
\newblock \bibinfo{journal}{CoRR} \URLprefix
  \url{https://arxiv.org/abs/1412.6621}.
\bibitem[{Rafegas and Vanrell(2017)}]{rafegas2017}
\bibinfo{author}{Rafegas, I.}, \bibinfo{author}{Vanrell, M.},
  \bibinfo{year}{2017}.
\newblock \bibinfo{title}{Color representation in cnns: parallelisms with
  biological vision}, in: \bibinfo{booktitle}{Int. Conference on Computer
  Vision (Workshops)}.
\bibitem[{Rafegas and Vanrell(2018)}]{rafegas2018}
\bibinfo{author}{Rafegas, I.}, \bibinfo{author}{Vanrell, M.},
  \bibinfo{year}{2018}.
\newblock \bibinfo{title}{Color encoding in biologically inspired convolutional
  neural networks}.
\newblock \bibinfo{journal}{Vision Research} \bibinfo{volume}{151},
  \bibinfo{pages}{7--17}.
\bibitem[{Shapley and Hawken(2011)}]{Shapley11}
\bibinfo{author}{Shapley, R.}, \bibinfo{author}{Hawken, M.},
  \bibinfo{year}{2011}.
\newblock \bibinfo{title}{Color in the cortex: Single- and double-opponent
  cells}.
\newblock \bibinfo{journal}{Vision Research} \bibinfo{volume}{51},
  \bibinfo{pages}{701--717}.
\bibitem[{Simonyan et~al.(2014)Simonyan, Vedaldi and Zisserman}]{Simonyan14}
\bibinfo{author}{Simonyan, K.}, \bibinfo{author}{Vedaldi, A.},
  \bibinfo{author}{Zisserman, A.}, \bibinfo{year}{2014}.
\newblock \bibinfo{title}{Deep inside convolutional networks: Visualising image
  classification models and saliency maps}, in: \bibinfo{booktitle}{Int. Conf.
  on Learning Representations}.
\bibitem[{Vedaldi and Lenc(2015)}]{vedaldi15matconvnet}
\bibinfo{author}{Vedaldi, A.}, \bibinfo{author}{Lenc, K.},
  \bibinfo{year}{2015}.
\newblock \bibinfo{title}{Matconvnet -- convolutional neural networks for
  matlab}, in: \bibinfo{booktitle}{Proceeding of the {ACM} Int. Conf. on
  Multimedia}.
\bibitem[{Yosinski et~al.(2015)Yosinski, Clune, Nguyen, Fuchs and
  Lipson}]{Yosinski15}
\bibinfo{author}{Yosinski, J.}, \bibinfo{author}{Clune, J.},
  \bibinfo{author}{Nguyen, A.}, \bibinfo{author}{Fuchs, T.},
  \bibinfo{author}{Lipson, H.}, \bibinfo{year}{2015}.
\newblock \bibinfo{title}{Understanding neural networks through deep
  visualization}, in: \bibinfo{booktitle}{Int. Conf. on ML}.
\bibitem[{Zeiler and Fergus(2014)}]{Zeiler14}
\bibinfo{author}{Zeiler, M.D.}, \bibinfo{author}{Fergus, R.},
  \bibinfo{year}{2014}.
\newblock \bibinfo{title}{Visualizing and understanding convolutional
  networks}, in: \bibinfo{booktitle}{In Proc. of ECCV}.
\bibitem[{Zeiler et~al.(2010)Zeiler, Krishnan, Taylor and Fergus}]{Zeiler10}
\bibinfo{author}{Zeiler, M.D.}, \bibinfo{author}{Krishnan, D.},
  \bibinfo{author}{Taylor, G.W.}, \bibinfo{author}{Fergus, R.},
  \bibinfo{year}{2010}.
\newblock \bibinfo{title}{Deconvolutional networks}, in:
  \bibinfo{booktitle}{Proc. CVPR}.
\bibitem[{{Zhou} et~al.(2018){Zhou}, {Bau}, {Oliva} and
  {Torralba}}]{Torralba2018_article}
\bibinfo{author}{{Zhou}, B.}, \bibinfo{author}{{Bau}, D.},
  \bibinfo{author}{{Oliva}, A.}, \bibinfo{author}{{Torralba}, A.},
  \bibinfo{year}{2018}.
\newblock \bibinfo{title}{Interpreting deep visual representations via network
  dissection}.
\newblock \bibinfo{journal}{IEEE Trans. on PAMI} .
\bibitem[{Zhou et~al.(2018)Zhou, Sun, Bau and Torralba}]{torralba_ablation}
\bibinfo{author}{Zhou, B.}, \bibinfo{author}{Sun, Y.}, \bibinfo{author}{Bau,
  D.}, \bibinfo{author}{Torralba, A.}, \bibinfo{year}{2018}.
\newblock \bibinfo{title}{Revisiting the importance of individual units in cnns
  via ablation}.
\newblock \bibinfo{journal}{CoRR} \URLprefix
  \url{https://arxiv.org/abs/1806.02891}.

\end{thebibliography}

\end{document}